\documentclass[conference]{IEEEtran}
\IEEEoverridecommandlockouts
\usepackage{cite}
\usepackage{amsmath,amssymb,amsfonts}
\usepackage{algorithm,algorithmic}

\usepackage{graphicx}
\usepackage{textcomp}
\usepackage{xcolor}
\usepackage{booktabs}
\usepackage{multirow}
\usepackage{etoolbox}
\usepackage{color}
\usepackage{ulem}
\usepackage{subfigure}
\makeatletter
\patchcmd{\@makecaption}
  {\scshape}
  {}
  {}
  {}
\makeatletter
\patchcmd{\@makecaption}
  {\\}
  {.\ }
  {}
  {}
\makeatother
\def\tablename{Table}
\def\BibTeX{{\rm B\kern-.05em{\sc i\kern-.025em b}\kern-.08em
    T\kern-.1667em\lower.7ex\hbox{E}\kern-.125emX}}
\begin{document}

\title{TREATED:~Towards Universal Defense against Textual Adversarial Attacks\\
}

\author{\IEEEauthorblockN{1\textsuperscript{st} Bin Zhu}
\IEEEauthorblockA{\textit{Cyberspace Institute of Advanced }\\
            \textit{Technology} \\
\textit{Guangzhou University}\\
Guangzhou, China \\
zhubin@e.gzhu.edu.cn}
\and
\IEEEauthorblockN{2\textsuperscript{nd} Zhaoquan Gu}
\IEEEauthorblockA{\textit{Cyberspace Institute of Advanced }\\
            \textit{Technology} \\
\textit{Guangzhou University}\\
Guangzhou, China \\
zqgu@gzhu.edu.cn}
\and
\IEEEauthorblockN{3\textsuperscript{th} Le Wang}
\IEEEauthorblockA{\textit{Cyberspace Institute of Advanced }\\
            \textit{Technology} \\
\textit{Guangzhou University}\\
Guangzhou, China \\
wangle@gzhu.edu.cn}
\and
\IEEEauthorblockN{4\textsuperscript{th} Zhihong Tian}
\IEEEauthorblockA{\textit{Cyberspace Institute of Advanced }\\
            \textit{Technology} \\
\textit{Guangzhou University}\\
Guangzhou, China \\
tianzhihong@gzhu.edu.cn}
}

\maketitle

\begin{abstract}

Recent work shows that deep neural networks are vulnerable to adversarial examples. Much work studies adversarial example generation, while very little work focuses on more critical adversarial defense. Existing adversarial detection methods usually make assumptions about the adversarial example and attack method (e.g., the word frequency of the adversarial example, the perturbation level of the attack method). However, this limits the applicability of the detection method. To this end, we propose TREATED, a universal adversarial detection method that can defend against attacks of various perturbation levels without making any assumptions. TREATED identifies adversarial examples through a set of well-designed reference models. Extensive experiments on three competitive neural networks and two widely used datasets show that our method achieves better detection performance than baselines. We finally conduct ablation studies to verify the effectiveness of our method.

\end{abstract}

\begin{IEEEkeywords}
Deep learning, Natural Language Processing, Adversarial Examples, Adversarial Detection
\end{IEEEkeywords}

\section{Introduction}
Through powerful expressive abilities, deep neural networks(DNNs) have achieved great success in many areas
\cite{kim-2014-convolutional, devlin-etal-2019-bert}.
However, recent work has revealed the fragility of DNNs. They are vulnerable to adversarial examples
\cite{szegedy-etal-2014-intriguing,goodfellow-etal-explaining}.
By adding invisible perturbations to the original examples, attackers can create adversarial examples which render the input misclassified. An increasing number of studies have shown that adversarial examples exist widely in computer vision(CV), natural language processing(NLP) and other domains. The existence of adversarial examples threatens the security-critical models and has garnered widespread attention\cite{Evtimov-etal-2017-robust,zhang-etal-2020-adversarial-attacks}.

In NLP, most work focuses on the generation of natural language adversarial examples to understand their nature
\cite{ebrahimi-etal-2018-hotflip,ren-etal-2019-generating,zang-etal-2020-word,jin-etal-2020-is,li-etal-2020-bert-attack}.
On the other hand, how to defend against textual adversarial attacks is a relatively more important and challenging problem. 
Existing defensive methods can be divided into two categories, \textit{adversarial training} and \textit{adversarial detection}.

Adversarial training refers to mixing adversarial examples with clean examples and retraining the model to improve the model’s robustness\cite{miyato-etal-2017-adversarial-training,zhu-etal-2020-freelb}. 
Adversarial training is widely used, but it is resource-consuming to retrain the model. Moreover, such defence methods make strong assumptions about attack methods(e.g., the strategy of word substitution, the perturbation level and so on), which limit their defensive capabilities and scope of application\cite{sato-etal-2018-interpretable,dong-etal-2021-towards-robustness}. 

Adversarial detection typically uses auxiliary components to recover or reject adversarial examples in the input to defend against adversarial attacks\cite{zhou-etal-2019-learning-discriminate,mozes-etal-2021-frequency}. In adversarial detection, retraining is not required. But the defender needs to make assumptions about the characteristic of adversarial examples(e.g., the word frequency) so that they can distinguish adversarial and clean examples. Since adversarial examples have not been thoroughly studied, these assumptions make adversarial detection more empirical. 

In this paper, we do not directly make assumptions about attack methods or adversarial examples but use a carefully constructed set of reference models to reflect the nature of adversarial examples indirectly. To be specific, we use the different predictions of reference models on clean and adversarial examples to distinguish them accurately and block adversarial attacks. 

Through empirical analysis, we argue that these reference models need to meet two criteria, 1.their predictions are consistent on clean examples, and 2.their predictions are inconsistent on the adversarial examples against the victim model. 

Through theoretical analysis, we prove the upper and lower bounds of detection accuracy that our method can achieve.

In practice, by decomposing the embeddings of the victim model, we obtain reference models that meet the above conditions to block adversarial attacks. 

Our contributions are summarized as follows:

\begin{itemize}
\item[1)] We propose TREATED, a universal method for blocking textual adversarial attacks without any assumption about attack methods or adversarial examples. Thus it can be applied in more realistic and complicated environments.
\item[2)] We empirically and theoretically illustrate the superiority of our method. Comprehensive experiments show that TREATED significantly outperforms two advanced defense methods. We will release our implementations for the convenience of future benchmarking.
\end{itemize}

\begin{figure*}
\begin{center}
\subfigure[]{
\includegraphics[width=0.7\columnwidth]{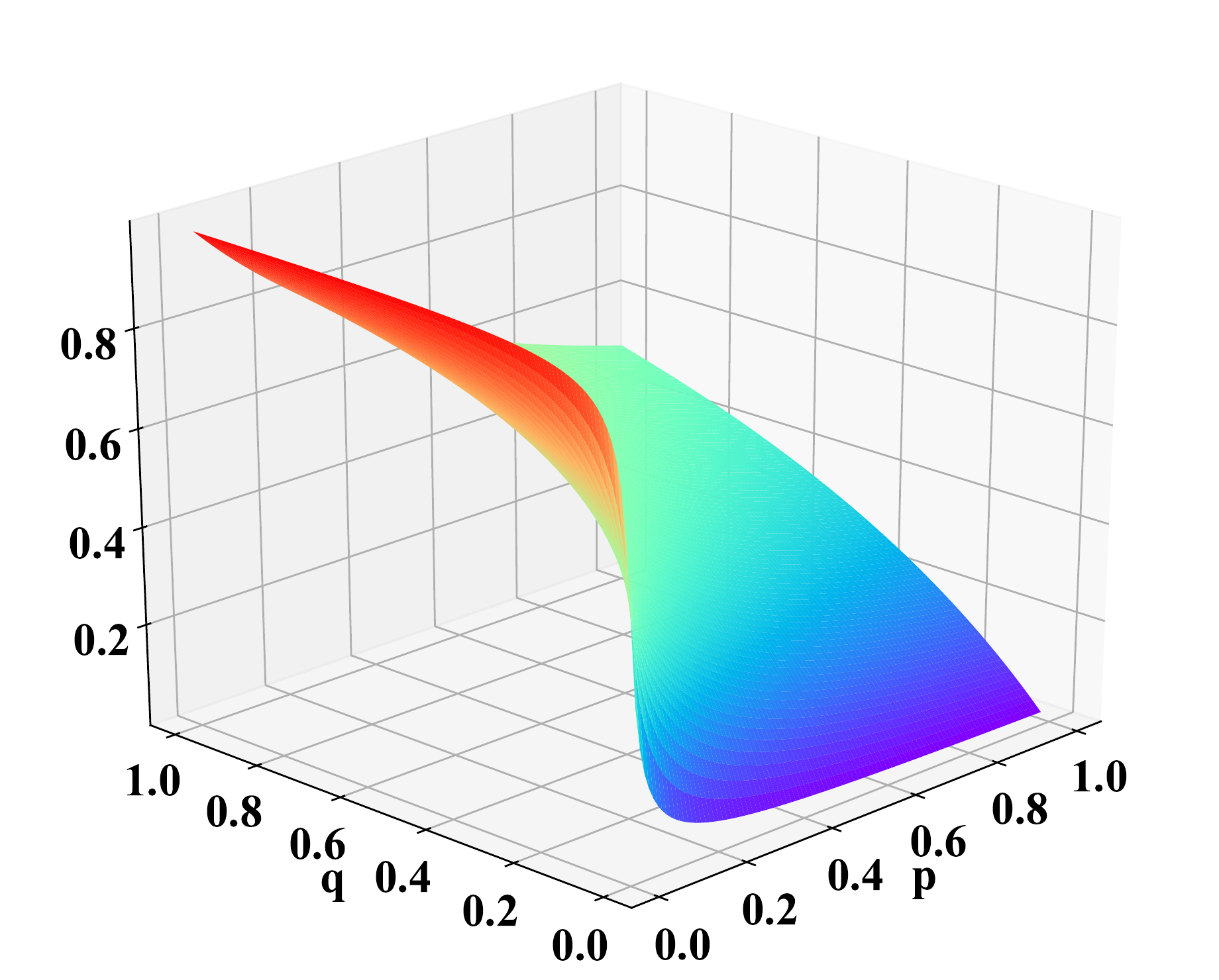}
}
\subfigure[]{
\includegraphics[width=0.75\columnwidth]{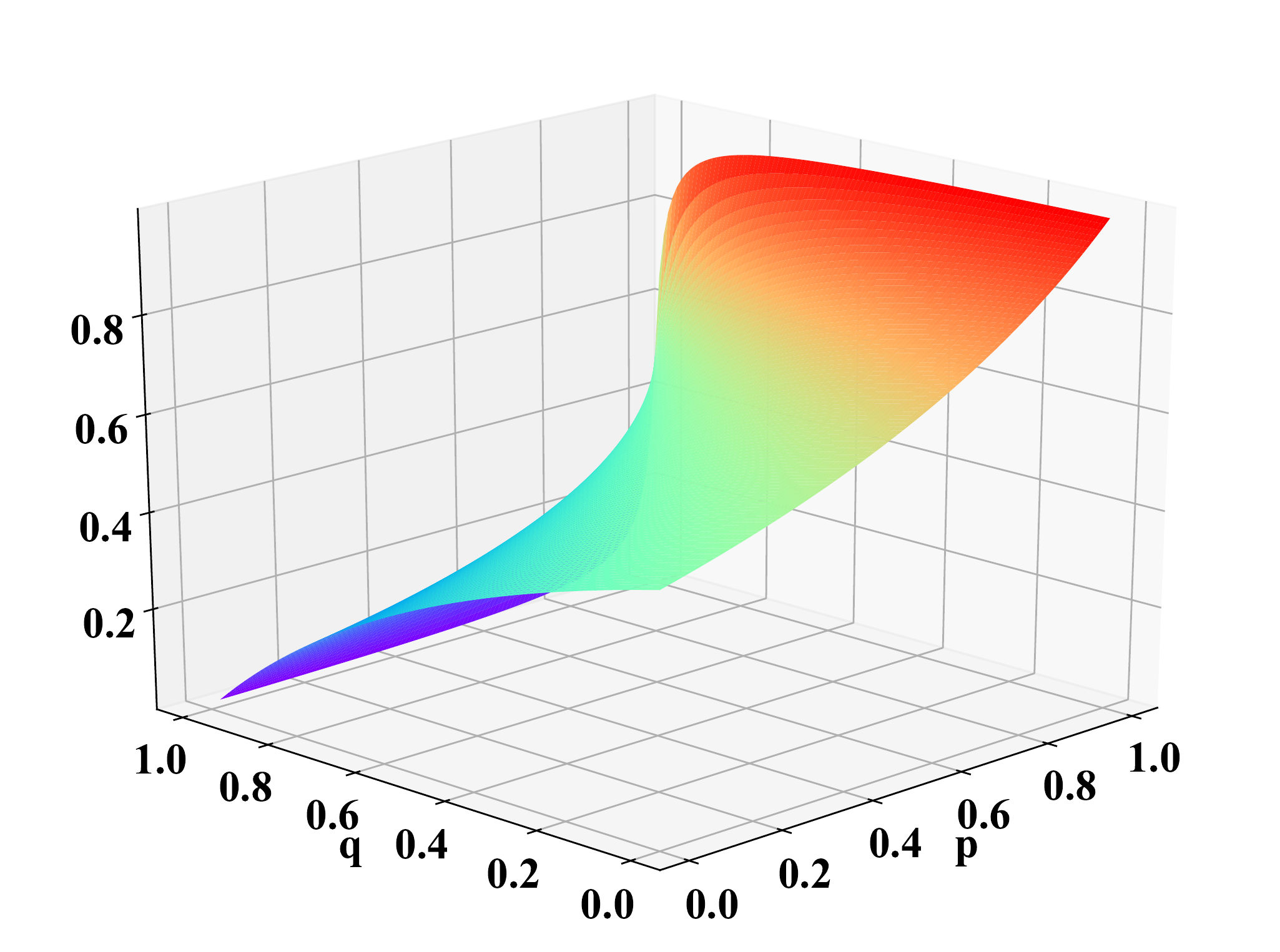}
}

\end{center}
\caption{(a)~Under the prior of $C$, the probability that $X$ is an adversarial example at different $p$ and $q$. (b)~Under the prior of $D$, the probability that $X$ is an adversarial example at different $p$ and $q$.}
\label{fig:p&q}
\end{figure*}


\section{Background and Related Work}

\subsection{Adversarial Attacks}

Given a classifier $F$ and an input-label pair $(x,y)$, the attacker aims to craft an adversarial example $x_{adv}$ misclassified by $F$ while $x_{adv}$ is in the $\epsilon$ neighborhood of $x$ as

\begin{equation}
\label{e1:adv.def}
    F(x_{adv})\neq y, s.t.~\left\|x-x_{adv}\right\| \leq \epsilon.
\end{equation}
Many works have leveraged the loss function of the classifier and the gradient ascent method to find qualified adversarial examples\cite{goodfellow-etal-explaining,Kurakin-etal-2017-adversarial,Madry-etal-2018-towards-deep}. However, this is not compatible with NLP tasks because the word embedding obtained by gradient information cannot correspond to a valid word. The valid words close to it can only be found by the projection method\cite{papernot-etal-2016-crafting,samanta-mehta-2017-towards}, which reduces the effectiveness of the attack.

Considering that the input of the NLP model is usually discrete tokens, the addition, deletion and replacement of tokens have naturally become common methods of textual adversarial attacks\cite{alzantot-etal-2018-generating,ren-etal-2019-generating}. For example, first, rank the tokens by importance and determine the replacement order; then use different strategies to find the best replacement for each token, thereby significantly reducing the time complexity of the search and forming an adversarial example. In addition, some works use paraphrasing\cite{iyyer-etal-2018-adversarial,zhang-etal-2019-paws}, text generation\cite{wang-etal-2020-cat,wang-etal-2020-t3}, generative adversarial networks(GAN)\cite{zhao-etal-2018-generating-natural}, reinforcement learning\cite{vijayaraghavan-etal-2019-generating} and other methods\cite{li-etal-2019-textbugger} to generate adversarial examples.

    

\subsection{Adversarial Defense}


\noindent\textbf{Adversarial Training}. Adversarial training is widely used to improve the robustness of DNNs\cite{goodfellow-etal-explaining,shafahi-etal-2019-adversarial,zhang-etal-2019-you-only}. By adding perturbations in the gradient direction to the training samples, adversarial training has achieved great success in many tasks. Recent studies have shown that adversarial training also helps to improve the generalization ability of DNNs\cite{zhu-etal-2020-freelb,jiang-etal-2020-smart}.

However, compared to standard training, adversarial training usually takes several times longer because the gradient of the training samples is repeatedly calculated, which is unacceptable in resource-constrained scenarios.


\noindent\textbf{Adversarial Detection}. The goal of adversarial detection is to identify the existence of adversarial examples, thereby rejecting them.
Most work aims to learn a representation to discriminate adversarial and benign inputs from the victim model\cite{li-li-2017-adversarial,grosse-etal-2017-on-the,ma-etal-2018-characterizing,feinman-etal-2017-detecting,metzen-etal-2017-detecting}.
However, in NLP, whether the examples are adversarial is challenging to judge, so some methods try to recover all inputs\cite{zhou-etal-2019-learning-discriminate,li-etal-2020-textshield}.
Nevertheless, this will inevitably destroy the original inputs and may cause the model's performance to decrease.

\noindent\textbf{Spelling Correction and Grammar Error Correction}. Spelling correction\cite{mays-etal-1991-context,zhang-etal-2020-spelling} and grammar error correction\cite{wang-etal-2020-comprehensive,sakaguchi-etal-2017-grammatical} are also used for blocking textual adversarial attacks. 
However, these methods can only deal with attacks that bring grammatical and spelling errors and cannot identify adversarial examples crafted by word substitution.

\section{TREATED}
\label{sec:3}

\subsection{Empirical Analysis}

Given a perfect classify $F_p$ which always classifies correctly, and the victim model $F_v$, we can always identify whether $F_v$ classifies correctly on input $X$(whether adversarial or not), blocking adversarial attacks because $F_p$ can always give the ground truth label $y_{true}$ of $X$. 

Note that $y_{true}$ is not necessary here. We only need to know whether the predictions of $F_p$ and $F_v$ are consistent. Then we can identify whether the prediction of $F_v$ is correct without knowing the ground truth label. It means that as long as there exists a set of reference models $\{F_r\}$ so that $F_{r_i}$ have consistent predictions on clean examples and inconsistent predictions on adversarial examples against $F_v$, we can identify whether $X$ is adversarial.

\subsection{Theoretical Analysis}
Mark event $A$ as $\{F_r\}$ have consistent predictions on clean examples, event $B$ as $\{F_r\}$ have consistent predictions on adversarial examples against $F_v$. We assume that $P(A)=p$ and $P(B)=q$. For an input $X$, mark event $C$($D$) as $\{F_r\}$ have consistent(inconsistent) predictions on $X$ and event $E$ as $X$ is adversarial. We have
\begin{equation}
\label{e2:p&q}
\begin{aligned}
    &P(E|C)=\frac{P(B)}{P(A)+P(B)}=\frac{q}{p+q},\\ &P(E|D)=\frac{1-P(B)}{(1-P(A))+(1-P(B))}=\frac{1-q}{2-(p+q)}.
\end{aligned}
\end{equation}

In our case, we expect $P(E|C)$ to be as small as possible and $P(E|D)$ to be as large as possible so that we can identify if $X$ is adversarial. For example, if $p=0.95$ and $q=0.1$, then $P(E|C)\approx0.095$ and $P(E|D)\approx0.947$. It means that we have a very high probability of making a correct judgment on whether $X$ is an adversarial example. Figure.\ref{fig:p&q} shows the probability of X is adversarial under prior $C$ and $D$ at different $p$ and $q$.

\subsection{Design of Reference Models}

We expect the reference models $\{F_r\}$ have a high $p$ and a low $q$ so that we can identify if $X$ is adversarial with a high accuracy. 

However, these two criteria seem to be contradictory. A high $p$ means that the models have similar parameter distributions and share a lot of knowledge, which also shares vulnerabilities, significantly enhancing the transferability of adversarial examples. In other words, adversarial examples can reduce the accuracy of $\{F_r\}$ to a low level so that the outputs of $\{F_r\}$ on the adversarial examples are also consistent, which hinders the detection of adversarial examples. Conversely, a low $q$ means that there are significant differences among $\{F_r\}$, and it is not easy to ensure consistent predictions on clean examples. So the key lies in how to construct $\{F_r\}$.

Recall the goal of textual adversarial attacks. It aims to find suitable perturbations and apply them to the original input. Essentially, this changes the embeddings that are finally input to the model. It inspires us to decompose the embedding layer of $F_v$ into $N$ parts and share the subsequent layers to construct $N$ reference models(i.e., $\{F_r\}$). In this way, we argue that $\{F_r\}$ have a very high $p$ since they are jointly trained. Moreover, adversarial examples act on the embedding layer of $F_v$, which is decomposed to construct $\{F_r\}$. Therefore $\{F_r\}$ share the increased loss caused by adversarial examples, thereby reducing their transferability, and thus have a low $q$. 

\textit{\textbf{Note that the same accuracy does not mean consistent predictions. Models may make different predictions on different samples but have the same accuracy.}}


\section{Experiments}

We attack three text classification models on two popular datasets using TextAttack\cite{morris-etal-2020-textattack-2}.  We also  make  reliable  evaluations  by automatic  evaluation.

\subsection{Experimental Setup}

\subsubsection{Baselines}

We choose FGWS\cite{mozes-etal-2021-frequency} as defense baselines since it achieves state-of-the-art performance in textual adversarial detection.
Follow \cite{zhou-etal-2019-learning-discriminate,mozes-etal-2021-frequency}, we use two word level attack methods, PWWS\cite{ren-etal-2019-generating} and Genetic\cite{alzantot-etal-2018-generating}, to make a fair comparison. 

\noindent \textbf{PWWS} generated adversarial texts against classification models by two strategies: 1) the word substitution strategy, and 2) the replacement order strategy. The word substitution strategy used the synonym substitution method. Through the named entity analysis techniques, it could improve the semantic similarity and reduce grammatical errors.

\noindent \textbf{Genetic} utilize the population-based optimization algorithm to reorganize the original sentence, yielding sentences with good attack effects.

Since TREATED is a universal defense method that can defend against attacks of  multiple perturbation levels, we further utilize TextBugger\cite{li-etal-2019-textbugger},  DeepWordBug\cite{gao-etal-2018-black} and TextFooler\cite{jin-etal-2020-is} to evaluate its defensive ability against attacks of different perturbation levels.

\subsubsection{Victim Models and Datasets}

Follow \cite{zhou-etal-2019-learning-discriminate,mozes-etal-2021-frequency}, we conduct experiments on two classification datasets. 

\begin{table}[ht]
    
    \centering
    \normalsize
    \caption{Summary of the datasets}
    \begin{tabular}{@{}ccccc@{}}
    \toprule
    \textbf{Dataset}     & \textbf{Classes} & \textbf{Train}   & \textbf{Test}   & \textbf{Avg Len} \\ \midrule
    SST-2       & 2       & 67,349   & 872  & 17          \\ 
    IMDb        & 2       & 25,000  & 25,000 & 201         \\ \bottomrule
    \end{tabular}
    
\label{t1_dataset}
\end{table}

\begin{table}[ht]
\centering
\normalsize
\caption{Parameters of the models.}
\begin{tabular}{@{}cccc@{}}
\toprule
               & \textbf{CNN} & \textbf{LSTM} & \textbf{RoBERTa} \\ \midrule
Embedding dim. & 300 & 300  & 768     \\
Filters        & 128 & -    & -       \\
Kernel size    & 3   & -    & -       \\
LSTM units     & -   & 128  & -       \\
Hidden dim.    & 100 & 100  & 100     \\ \bottomrule
\end{tabular}
\label{t:model_para}
\end{table}

\noindent\textbf{The Stanford Sentiment Treebank(SST-2)}\cite{socher-etal-2013-recursive} is a sentence-level sentiment analysis dataset. The movie reviews are
given by professionals.

\noindent \textbf{The IMDb reviews dataset}\cite{maas-etal-2011-learning} is a document-level sentiment analysis dataset containing 50,000
non-professional movie reviews, of which 25,000 are positive, and 25,000
are negative reviews.

We perform adversarial attacks on three neural networks, WordCNN\cite{kim-2014-convolutional}, LSTM\cite{hochreiter-schmidhuber-1997-long} and RoBERTa\cite{liu-etal-2019-roberta}.

\begin{table*}[ht]
\centering
\normalsize
\caption{Adversarial detection performances of FGWS and TREATED, including the accuracy increased after adversarial detection(Increased acc.), the true positive rate and false positive rate(TPR, FPR) and F1.}
\begin{tabular}{@{}llcccccc@{}}
\toprule
\multicolumn{1}{c}{\multirow{2}{*}{\textbf{Dataset/Model}}} &
  \multicolumn{1}{c}{\multirow{2}{*}{\textbf{Attack}}} &
  \multicolumn{2}{c}{\textbf{Increased acc.}} & 
  \multicolumn{2}{c}{\textbf{TPR(FPR)}} &
  \multicolumn{2}{c}{\textbf{F1}} \\ \cmidrule(l){3-4} \cmidrule(l){5-6} \cmidrule(l){7-8} 
\multicolumn{1}{c}{} &
  \multicolumn{1}{c}{} &
  FGWS &
  TREATED &
  FGWS &
  TREATED &
  FGWS &
  TREATED \\ \midrule
\multirow{2}{*}{SST-2/LSTM} & PWWS      & +30.0 & \textbf{+35.7}  & - & 62.2(20.49)  & 63.9 & \textbf{68.1} \\
                            & Genetic   & +29.2 & \textbf{+33.5}  & - & 61.0(21.47)  & 60.3 & \textbf{65.0} \\ \cmidrule(l){3-4} \cmidrule(l){5-6} \cmidrule(l){7-8} 
\multirow{2}{*}{IMDb/CNN}  & PWWS      & +60.0 & \textbf{+84.8}  & - & 91.9(0.05)  & 83.9 & \textbf{93.2} \\
                           & Genetic   & +57.8 & \textbf{+66.3}  & - & 90.5(0.14)  & 83.5 & \textbf{88.4} \\ \bottomrule
\end{tabular}%
\label{t2:benchmark_cnn_lstm}
\end{table*}

Both the datasets and networks are widely used in adversarial learning in NLP. Details can be found in Table.~\ref{t1_dataset} and Table.~\ref{t:model_para}.

\subsection{Experimental Results}

We generate 1,000 adversarial examples for each model, dataset, and attack method as part of the detection set. The corresponding clean examples are used as the other part of the detection set. We report the increased accuracy of the model after adversarial detection and the true positive rate(TPR), false positive rate(FPR) and F1 of the detection method to make a comprehensive evaluation.

We make a detailed comparison with FGWS, the state-of-the-art detecting method. Table.~\ref{t2:benchmark_cnn_lstm} shows the overall defense performance on CNN and LSTM models. As can be seen, TREATED can restore the models to higher accuracy. The defense performance decline on the SST-2 dataset. But it still outperforms FGWS. Our method performs better in detection accuracy, improving the F1 score by up to 9.3\% on the CNN model and IMDb dataset.

We report the results of defending against more attacks on the RoBERTa model in Table.\ref{t3:benchmark_r}.
TREATED performs well against attacks of different perturbation levels, such as word-level, char-level and multi-level. It can always increase the accuracy of the models by more than 40\% in the face of various attacks. 

\begin{table}[ht]
\centering
\normalsize
\caption{Defense performances on IMDb dataset and RoBERTa against more attacks.}
\begin{tabular}{@{}lccc@{}}
\toprule
\textbf{Attack}      & \textbf{Increased acc.} & \textbf{TPR(FPR)} & \textbf{F1} \\ \midrule
PWWS        &   +64.7       & 69.4(14.47)  &  75.5  \\
Genetic     &   +40.0       & 71.4(35.71)  &  69.0  \\
DeepWordBug &   +40.4       & 58.6(16.11)  &  67.1  \\
TextBugger  &   +45.0       & 49.9(12.65)  &  61.4  \\ \bottomrule
\end{tabular}%
\label{t3:benchmark_r}
\end{table}

\section{Discussion}

\subsection{Influence on Unperturbed Data}

We further study the effect of TREATED on the original accuracy of the models. We use the original test set as the detection set to observe whether the accuracy decrease significantly.

As can be seen in Table.~\ref{t6:unpert}, the accuracy of WordCNN and RoBERTa drop by 3.5\% and 2.6\%, respectively, which are almost negligible. It shows that TREATED can block adversarial attacks without affecting the original performance of the model.

\begin{table}[ht]
\centering
\normalsize
\caption{Influence of TREATED on unperturbed data.}
\begin{tabular}{@{}ccc@{}}
\toprule
Dataset/Model & Orinigal acc. & Detecting acc. \\ \midrule
IMDb/CNN      & 89.3          & 85.8           \\
IMDb/RoBERTa  & 90.9          & 88.3           \\ \bottomrule
\end{tabular}%
\label{t6:unpert}
\end{table}

\subsection{Ablation Study}

We construct $\{F_r\}$ by decomposing the embedding layer of the victim model. In order to illustrate its effectiveness, we conduct thorough ablation studies. We use standard training models(STM) as reference models to detect adversarial examples.

Table.~\ref{t4:ablation_levels} shows the detection performance of STM and TREATED. Under word- and char-level attacks, TREATED can increase the accuracy of victim models by more than 88\%, which means that TREATED completely prevents adversarial attacks. On the most difficult multi-level to defend, treated increases the accuracy of the victim model by 77.4\%. TREATED also maintains a TPR above 96\%, an FPR below 6\% and an F1 above 95\%, indicating very few misjudgments.

Then we replace the reference models with STM. The defense performance significantly decreases, which fully illustrates the superiority of decomposition of the embedded layer.

In order to further verify the advantages of decomposition of the embedding layer, we report $p$ and $q$ under different conditions. As shown in table~\ref{t5:ablation_pq}, our method always has a higher $p$ and a lower $q$, verifying that our method will have a higher detection accuracy.
\begin{table*}[ht]
\centering
\normalsize
\caption{Ablation study on adversarial detection performances for TREATED and STM against attacks of different perturbation levels. The victim model is WordCNN trained on IMDb.}
\begin{tabular}{@{}lllclc@{}}
\toprule
\textbf{Attack}      & \textbf{Level} & \textbf{Defense} & \textbf{Increased acc.} & \textbf{TPR(FPR)} & \textbf{F1} \\ \midrule
\multirow{2}{*}{TextFooler}  & \multirow{2}{*}{word}   & STM      &  +58.7    & 63.6(15.82)        & 70.9   \\
                             &                         & TREATED  &  \textbf{+88.6}    & \textbf{96.0(5.20)}         & \textbf{95.4} \\ \cmidrule(l){3-6}
\multirow{2}{*}{DeepWordBug} & \multirow{2}{*}{char}   & STM      &  +68.6    & 70.5(18.16)        & 74.8    \\ 
                             &                         & TREATED  &  \textbf{+88.5}    & \textbf{96.2(5.22)}         & \textbf{95.5} \\ \cmidrule(l){3-6}
\multirow{2}{*}{TextBugger}  & \multirow{2}{*}{multi}  & STM      &  +61.2    & 66.2(15.87)        & 72.7   \\
                             &                         & TREATED  &  \textbf{+77.4}    & \textbf{96.3(5.97)}         & \textbf{95.2} \\
\bottomrule

\end{tabular}%
\label{t4:ablation_levels}
\end{table*}

\begin{table*}[ht]
\centering
\normalsize
\caption{Ablation study on $p$ and $q$ on different victim models and test sets between TREATED and STM. The attack method is PWWS.}
\begin{tabular}{@{}ccccccc@{}}
\toprule
 & \multicolumn{2}{c}{\textbf{SST-2/LSTM}} & \multicolumn{2}{c}{\textbf{IMDb/CNN}} & \multicolumn{2}{c}{\textbf{IMDb/RoBERTa}} \\ \cmidrule(l){2-3} \cmidrule(l){4-5} \cmidrule(l){6-7}
  & STM & TREATED & STM & TREATED & STM & TREATED \\ \cmidrule(l){1-3} \cmidrule(l){4-5} \cmidrule(l){6-7}
$p \uparrow$ & 0.7052    & \textbf{0.7951}        & 0.7842    & \textbf{0.8723}   & 0.8422    &  \textbf{0.8840}       \\
$q \downarrow$ & 0.6471    & \textbf{0.3777}       & 0.3142    & \textbf{0.0813}   & 0.5483    & \textbf{0.5006}         \\ \bottomrule
\end{tabular}%
\label{t5:ablation_pq}
\end{table*}


\section{Conclusion and Future Work}

We propose TREATED, a universal adversarial detection method to defend against textual adversarial attacks with multiple perturbation levels through empirical and theoretical analysis. 
By decomposing the embedding layer, we construct a set of reference models that can make consistent predictions on clean examples and inconsistent predictions on adversarial examples. Thus we can identify adversarial examples and block attacks via these reference models.
Extensive experiments prove the effectiveness of TREATED. Compared with the state-of-the-art defense method, our method has higher detection accuracy without harming the original performance. Ablation studies illustrate the superiority of our embedding decomposition method.

Further improving the defense performance on large-scale pre-training model(e.g., RoBERTa) is one of our future works. Besides, it is also promising to find other ways to construct $\{F_r\}$.

\normalem
\bibliographystyle{IEEEtran}
\bibliography{anthology,ref}

\end{document}